\documentclass[11pt]{article}

\usepackage[final]{acl}

\usepackage{times}
\usepackage{latexsym}
\usepackage{subcaption}

\usepackage[T1]{fontenc}

\usepackage[utf8]{inputenc}

\usepackage{microtype}
\usepackage{comment}

\usepackage{graphicx}
\usepackage[inline]{enumitem}

\usepackage{hyperref}

\let\ACLendabstract\endabstract

\usepackage{arabtex}
\usepackage{utf8}
\setcode{utf8}

\let\endabstract\ACLendabstract
\usepackage{booktabs}

\usepackage{xcolor}
\usepackage{tikz}
\usepackage{amsmath}

\usetikzlibrary{positioning, fit, backgrounds}

\definecolor{allagree}{RGB}{180,220,180}    
\definecolor{twoagree}{RGB}{255,220,130}    
\definecolor{disagree}{RGB}{255,170,160}    
\definecolor{negcol}{RGB}{220,220,220}      

\makeatletter
\protected\def\begin#1{%
  \UseHook{env/#1/before}%
  \@ifundefined{#1}%
    {\def\reserved@a{\@latex@error{Environment #1 undefined}\@eha}}%
    {\def\reserved@a{\def\@currenvir{#1}%
        \edef\@currenvline{\on@line}%
        \@execute@begin@hook{#1}%
        \csname #1\endcsname}}%
  \@ignorefalse
  \begingroup
  \let\end\a@l@end
  \@endpefalse\reserved@a}
\makeatother

\title{AraMIP: Extending MIPVU Towards Metaphor Identification in Arabic}

\author{
    Mandar Marathe\textsuperscript{*1} \:
    Manar Ali\textsuperscript{*2} \:
    Sara Nabhani\textsuperscript{*3} \:
    Raia Abu Ahmad\textsuperscript{*4,5} \\
    \textbf{Ibrahim Baroud\textsuperscript{*4,5}} \:
    \textbf{Omar Momen\textsuperscript{*2}} \\
    \textsuperscript{1}SOAS University of London, United Kingdom \\
    \textsuperscript{2}CRC 1646 `Linguistic Creativity in Communication', Bielefeld University, Germany \\
    \textsuperscript{3}Computational Linguistics, CLCG, University of Groningen, The Netherlands \\
    \textsuperscript{4}German Research Center for Artificial Intelligence (DFKI), Germany \\
    \textsuperscript{5}Quality \& Usability Lab, Technical University of Berlin, Germany \\
    \small
    \textsuperscript{1}\texttt{mandar@marathe.org}
    \hspace{4em}
    \textsuperscript{3}\texttt{s.nabhani@rug.nl} \\
    \small
    \textsuperscript{2}\texttt{\{manar.ali,omar.hassan\}@uni-bielefeld.de}
    \quad
    \textsuperscript{4,5}\texttt{\{raia.abu\_ahmad,ibrahim.baroud\}@dfki.de}
}

\begin{document}
\maketitle
\def\thefootnote{*}\footnotetext{Equal contribution.}\def\thefootnote{\arabic{footnote}}

\begin{abstract}
Metaphor research has gained increasing attention due to its relevance to linguistic creativity, language use, cognitive processes, and related areas. While many efforts have been devoted to metaphor identification and annotation in English and other languages, Arabic remains under-resourced in this area. In this work, we propose the Arabic Metaphor Identification Procedure (AraMIP), a novel guideline for Arabic metaphor annotation. AraMIP builds on the widely used Metaphor Identification Procedure Vrije Universiteit (MIPVU) framework, incorporating adaptations that accounts for the language-specific properties of Arabic. We distinguish three major types of Arabic figurative language: \textit{isti‘āra} (metaphor), \textit{kināya} (metonymy/indirect expression), and \textit{tashbīh} (simile) and annotate a pilot dataset of 300 sentences (5277 words). Our analysis reveals key challenges specific to Arabic, including morphological complexity, inconsistencies in dictionary sense ordering, and the absence of standardized contextual materials for annotators. This work contributes a first step toward standardized Arabic figurative instances and facilitates the development of larger annotated resources, thereby supporting future research on figurative language in Arabic.\footnote{Data, annotation guidelines and code are available here \url{https://github.com/AraMIP/AraMIP}}
\end{abstract}

\section{Introduction}

Humans use metaphors in everyday language to achieve diverse communicative goals, such as expressing emotions and explaining unfamiliar concepts. For instance, an L2 speaker may describe an attempted scam by saying, “Someone wants to eat my money,” drawing on the concrete domain of eating to express the more abstract notion of financial exploitation. Notably, humans can interpret metaphorical expressions quickly \citep{glucksberg1998understanding}. Such expressions involve cross-domain mapping, which is central to several influential theories of metaphor, most notably the Conceptual Metaphor Theory \cite[CMT,][]{lakoff1980metaphors}, which argues that a metaphor is not only a linguistic phenomenon but also a cognitive mechanism through which concrete concepts (source) are mapped onto more abstract domains (target).
In the example above, \textit{eating} is the source and \textit{financial exploitation} the target. Metaphorical framing also plays an important role in shaping how people conceptualize abstract ideas and reason about them \cite{thibodeau2011metaphors}.

Metaphor research has attracted increasing attention in (Cognitive) Linguistics and Natural Language Processing (NLP), and has also been discussed in the broader context of linguistic creativity \citep{vogel2026linguistic}. 
Data collection and annotation studies identified metaphors~\citep{mipvu,mohammad-etal-2016-metaphor,beigman-klebanov-etal-2018-corpus,leong-etal-2018-report}, or classified them based on deliberateness and novelty~\cite{dmip,do-dinh-etal-2018-weeding}.
~\citet{tong-etal-2021-recent} and ~\citet{ge2023survey} investigated language models' ability to generate and comprehend metaphors as a proxy for cognitive and pragmatic competence.
Other studies explored the role of metaphor in downstream tasks such as machine translation~\cite{wang-etal-2024-mmte} and offensive language detection~\cite{lemmens-etal-2021-improving,zeng-etal-2025-sheeps}, where identifying figurative expressions can improve system performance~\cite{huguet-cabot-etal-2020-pragmatics,dankers-etal-2019-modelling}.

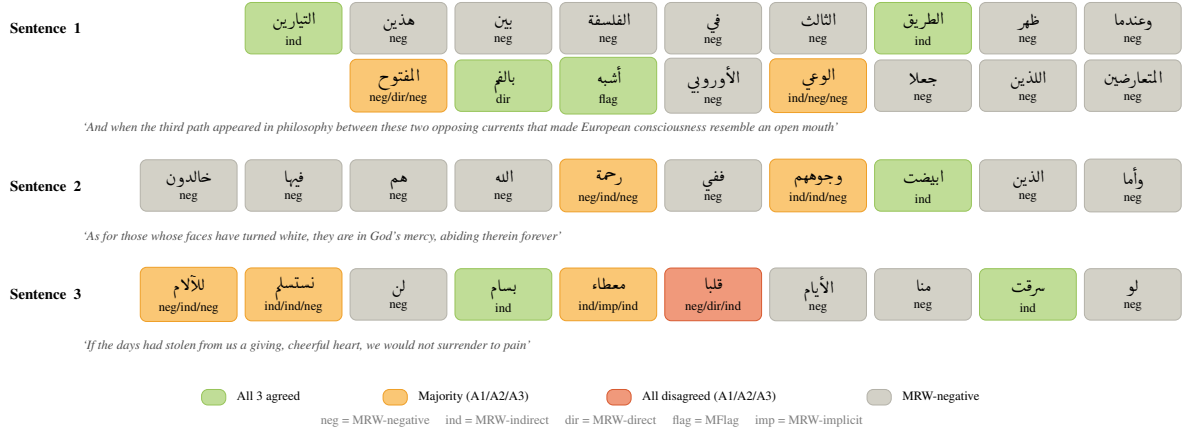
\begin{figure*}[h]
\centering
\scalebox{0.80}{
\begin{tikzpicture}[
  tok/.style={rectangle, rounded corners=3pt, draw=gray, fill=#1,
              minimum width=1.6cm, minimum height=0.85cm,
              align=center, font=\scriptsize},
  node distance=0.18cm
]

\definecolor{cagr}{RGB}{192,221,151}
\definecolor{bagr}{RGB}{151,196,89}
\definecolor{cmaj}{RGB}{250,199,117}
\definecolor{bmaj}{RGB}{239,159,39}
\definecolor{cdis}{RGB}{240,153,123}
\definecolor{bdis}{RGB}{216,90,48}
\definecolor{cneg}{RGB}{211,209,199}
\definecolor{bneg}{RGB}{180,178,169}

\node[font=\scriptsize\bfseries, text=black, anchor=west] at (0,0) {Sentence ~1};

\node[tok=cneg, draw=bneg, anchor=east] at (19.5, 0)    (s1t1)  {\RL{\<وعندما>}    \\ \tiny neg};
\node[tok=cneg, draw=bneg, left=0.12cm of s1t1]         (s1t2)  {\RL{\<ظهر>}       \\ \tiny neg};
\node[tok=cagr, draw=bagr, left=0.12cm of s1t2]         (s1t3)  {\RL{\<الطريق>}    \\ \tiny ind};
\node[tok=cneg, draw=bneg, left=0.12cm of s1t3]         (s1t4)  {\RL{\<الثالث>}    \\ \tiny neg};
\node[tok=cneg, draw=bneg, left=0.12cm of s1t4]         (s1t5)  {\RL{\<في>}        \\ \tiny neg};
\node[tok=cneg, draw=bneg, left=0.12cm of s1t5]         (s1t6)  {\RL{\<الفلسفة>}   \\ \tiny neg};
\node[tok=cneg, draw=bneg, left=0.12cm of s1t6]         (s1t7)  {\RL{\<بين>}       \\ \tiny neg};
\node[tok=cneg, draw=bneg, left=0.12cm of s1t7]         (s1t8)  {\RL{\<هذين>}      \\ \tiny neg};
\node[tok=cagr, draw=bagr, left=0.12cm of s1t8]         (s1t9)  {\RL{\<التيارين>}  \\ \tiny ind};

\node[tok=cneg, draw=bneg, anchor=east] at (19.5,-0.95) (s1t10) {\RL{\<المتعارضين>}\\ \tiny neg};
\node[tok=cneg, draw=bneg, left=0.12cm of s1t10]        (s1t11) {\RL{\<اللذين>}    \\ \tiny neg};
\node[tok=cneg, draw=bneg, left=0.12cm of s1t11]        (s1t12) {\RL{\<جعلا>}      \\ \tiny neg};
\node[tok=cmaj, draw=bmaj, left=0.12cm of s1t12]        (s1t13) {\RL{\<الوعي>}     \\ \tiny ind/neg/neg};
\node[tok=cneg, draw=bneg, left=0.12cm of s1t13]        (s1t14) {\RL{\<الأوروبي>}  \\ \tiny neg};
\node[tok=cagr, draw=bagr, left=0.12cm of s1t14]        (s1t15) {\RL{\<أشبه>}      \\ \tiny flag};
\node[tok=cagr, draw=bagr, left=0.12cm of s1t15]        (s1t16) {\RL{\<بالفم>}     \\ \tiny dir};
\node[tok=cmaj, draw=bmaj, left=0.12cm of s1t16]        (s1t17) {\RL{\<المفتوح>}   \\ \tiny neg/dir/neg};

\node[font=\tiny\itshape, text=black!70, anchor=west] at (1.2,-1.65)
  {`And when the third path appeared in philosophy between these two opposing currents that made European consciousness resemble an open mouth'};

\node[font=\scriptsize\bfseries, text=black, anchor=west] at (0,-2.6) {Sentence ~2};

\node[tok=cneg, draw=bneg, anchor=east] at (19.5,-2.6) (s2t1)  {\RL{\<وأما>}      \\ \tiny neg};
\node[tok=cneg, draw=bneg, left=0.12cm of s2t1]        (s2t2)  {\RL{\<الذين>}     \\ \tiny neg};
\node[tok=cagr, draw=bagr, left=0.12cm of s2t2]        (s2t3)  {\RL{\<ابيضت>}     \\ \tiny ind};
\node[tok=cmaj, draw=bmaj, left=0.12cm of s2t3]        (s2t4)  {\RL{\<وجوههم>}    \\ \tiny ind/ind/neg};
\node[tok=cneg, draw=bneg, left=0.12cm of s2t4]        (s2t5)  {\RL{\<ففي>}       \\ \tiny neg};
\node[tok=cmaj, draw=bmaj, left=0.12cm of s2t5]        (s2t6)  {\RL{\<رحمة>}      \\ \tiny neg/ind/neg};
\node[tok=cneg, draw=bneg, left=0.12cm of s2t6]        (s2t7)  {\RL{\<الله>}      \\ \tiny neg};
\node[tok=cneg, draw=bneg, left=0.12cm of s2t7]        (s2t8)  {\RL{\<هم>}        \\ \tiny neg};
\node[tok=cneg, draw=bneg, left=0.12cm of s2t8]        (s2t9)  {\RL{\<فيها>}      \\ \tiny neg};
\node[tok=cneg, draw=bneg, left=0.12cm of s2t9]        (s2t10) {\RL{\<خالدون>}    \\ \tiny neg};

\node[font=\tiny\itshape, text=black!70, anchor=west] at (1.2,-3.45)
  {`As for those whose faces have turned white, they are in God's mercy, abiding therein forever'};

\node[font=\scriptsize\bfseries, text=black, anchor=west] at (0,-4.4) {Sentence ~3};

\node[tok=cneg, draw=bneg, anchor=east] at (19.5,-4.4) (s3t1)  {\RL{\<لو>}        \\ \tiny neg};
\node[tok=cagr, draw=bagr, left=0.12cm of s3t1]        (s3t2)  {\RL{\<سرقت>}      \\ \tiny ind};
\node[tok=cneg, draw=bneg, left=0.12cm of s3t2]        (s3t3)  {\RL{\<منا>}       \\ \tiny neg};
\node[tok=cneg, draw=bneg, left=0.12cm of s3t3]        (s3t4)  {\RL{\<الأيام>}    \\ \tiny neg};
\node[tok=cdis, draw=bdis, left=0.12cm of s3t4]        (s3t5)  {\RL{\<قلبا>}      \\ \tiny neg/dir/ind};
\node[tok=cmaj, draw=bmaj, left=0.12cm of s3t5]        (s3t6)  {\RL{\<معطاء>}     \\ \tiny ind/imp/ind};
\node[tok=cagr, draw=bagr, left=0.12cm of s3t6]        (s3t7)  {\RL{\<بسام>}      \\ \tiny ind};
\node[tok=cneg, draw=bneg, left=0.12cm of s3t7]        (s3t8)  {\RL{\<لن>}        \\ \tiny neg};
\node[tok=cmaj, draw=bmaj, left=0.12cm of s3t8]        (s3t9)  {\RL{\<نستسلم>}    \\ \tiny ind/ind/neg};
\node[tok=cmaj, draw=bmaj, left=0.12cm of s3t9]        (s3t10) {\RL{\<للآلام>}    \\ \tiny neg/ind/neg};

\node[font=\tiny\itshape, text=black!70, anchor=west] at (1.2,-5.25)
  {`If the days had stolen from us a giving, cheerful heart, we would not surrender to pain'};

\node[tok=cagr, draw=bagr, minimum width=0.4cm, minimum height=0.25cm] at (3.5,-6.1) {};
\node[font=\tiny, anchor=west] at (3.75,-6.1) {All 3 agreed};
\node[tok=cmaj, draw=bmaj, minimum width=0.4cm, minimum height=0.25cm] at (6.5,-6.1) {};
\node[font=\tiny, anchor=west] at (6.75,-6.1) {Majority (A1/A2/A3)};
\node[tok=cdis, draw=bdis, minimum width=0.4cm, minimum height=0.25cm] at (10.2,-6.1) {};
\node[font=\tiny, anchor=west] at (10.45,-6.1) {All disagreed (A1/A2/A3)};
\node[tok=cneg, draw=bneg, minimum width=0.4cm, minimum height=0.25cm] at (14.5,-6.1) {};
\node[font=\tiny, anchor=west] at (14.75,-6.1) {MRW-negative};

\node[font=\tiny, text=gray, anchor=center] at (9.75,-6.5)
  {neg = MRW-negative \quad ind = MRW-indirect \quad dir = MRW-direct \quad flag = MFlag \quad imp = MRW-implicit};

\end{tikzpicture}
}
\caption{Three annotated sentences illustrating token-level agreement.
Green = all 3 agreed; orange = majority; red = all disagreed; gray = MRW-negative.
Disagreement shown as A1/A2/A3.}
\label{fig:annotation-examples}
\end{figure*}

Besides English, annotation efforts and benchmark datasets have been developed for several languages, including Spanish~\cite{sanchez-bayona-agerri-2022-leveraging}, Chinese~\cite{lu_towards_2017}, and German~\cite{egg-kordoni-2022-metaphor}. Arabic, however, remains under-resourced in this area. While several studies have touched on Arabic metaphor in NLP contexts~\cite{alkhatib2016natural, alsiyat-piao-2020-metaphorical, amc-corpus}, to our knowledge none provide a standardized, replicable annotation procedure, and existing datasets are limited in domain coverage or methodological transparency~\cite{alsiyat-piao-2020-metaphorical,magdy-etal-2024-gazelle}. Arabic furthermore presents challenges that existing frameworks do not address. In particular, its rich morphological system, root-based lexicographic structure, and the long rhetorical tradition distinguishing \textit{isti'āra} (metaphor), \textit{kināya} (metonymy/indirect expression), and \textit{tashbīh} (simile) in ways that do not map directly onto categories developed for other languages.

In this work, we introduce the Arabic Metaphor Identification Procedure (AraMIP), a two-stage annotation framework that adapts the widely used Metaphor Identification Procedure Vrije Universiteit~\cite[MIPVU,][]{mipvu} to Arabic. The first stage follows MIPVU's word-level metaphor identification (see Figure~\ref{fig:annotation-examples}); the second maps metaphor-related words (spans) onto Arabic rhetorical categories and records the conventionality of each construction on a five-point scale. We apply AraMIP to a pilot corpus of 300 sentences (5277 words) sampled from the BAREC-10M corpus~\cite{elmadani-etal-2026-large} and release the guidelines and the annotations to support future work. Our pilot study surfaces concrete challenges specific to Arabic annotation, including dictionary coverage at the boundary between Modern Standard Arabic (MSA) and Classical Arabic, and root-to-derivative sense resolution, and provides a diagnostic roadmap for future iterations of the framework.

\section{Background and Related Work }


\subsection{Metaphor Theories}

Theories and definitions of metaphors have a very long history. We do not review all theories of metaphors, but we point out their major views and understandings. The oldest standard view comes from \citeauthor{aristotle1996poetics}, where metaphor was mainly seen as a decorative, poetic, or rhetorical device where one word is used instead of another. \citet{black1955metaphor} viewed that metaphorical meaning emerges from the interaction between two conceptual systems, not by simple substitution as in the classical theory. On the other hand, in pragmatics, metaphor is not primarily a special semantic mapping. It is a case where the meaning of the sentence and the meaning of the speaker separate~\citep{searle1979metaphor}. Recently, CMT that was posed by \citet{lakoff1980metaphors} has become widely adopted in multiple disciplines. It states that metaphor arises from understanding one domain of experience (that is typically abstract) in terms of another (that is typically concrete), e.g., in ``She attacked my point.,'' the abstract target domain of ``Argument'' is understood in terms of the source concrete domain of ``War.'' 

In Arabic literary studies, particularly in the field of \textit{al-balāgha}\footnote{al-balāgha (Arabic rhetoric) is the classical Arabic tradition concerned with effective, expressive, and aesthetically appropriate uses of language.}, metaphor is discussed within the domain of \textit{‘ilm al-bayān}\footnote{‘ilm al-bayān (the science of figurative expression) is a branch of classical Arabic rhetoric concerned with the ways a meaning can be expressed figuratively or indirectly}, especially through the interrelated concepts of \textit{tashbīh} (simile), \textit{isti‘āra} (metaphor), and \textit{kināya} (metonymy)~\cite{abdulraof2006arabic}. The dominant textbook definition derives \textit{isti‘āra} from \textit{tashbīh}, where one pole of an underlying comparison is omitted. A more sophisticated account of metaphor is provided by Abd al-Qāhir al-Jurjānī (d. 1078) as explained in~\citet{larkin1995theology}, who views metaphor not merely as an ornamental substitution, but as part of the production of second-order meaning through structure, context, and imaginative interpretation.

\subsection{Operationalizing Metaphors}

A group of scholars~\citep{mip2007} worked on developing a standard procedure for identifying metaphors in written and spoken discourse. They introduced the Metaphor Identification Procedure (MIP), which was then extended with finer details to MIPVU~\citep{mipvu}.\footnote{See the MIPVU procedure in Appendix~\ref{sec: apx_mipvu}.} MIP was motivated by the fact that scholars often disagreed about what counts as metaphor. The authors aimed for an explicit, reliable, and flexible method for identifying metaphorically used words in written and spoken discourse. The core theoretical assumption of MIP is that a lexical unit (a word) is metaphorically used when its contextual meaning contrasts with a more basic meaning, but the contextual meaning can be understood in comparison with that basic meaning. 

The operationalization framework of MIP resulted in a standard annotated metaphor corpus in English~\cite[VUAMC,][]{mipvu}, which has served as the standard corpus for most metaphor studies in NLP since then~\citep{leong-etal-2018-report,leong-etal-2020-report}. Later, adaptations of MIP to other languages were developed and resulted in metaphor corpora in languages other than English. For example: Russian~\citep{badryzlova_annotating_2013}, Chinese~\citep{lu_towards_2017}, German~\citep{egg-kordoni-2022-metaphor}, Spanish~\citep{sanchez-bayona-agerri-2022-leveraging} and Japanese~\cite{zhu-etal-2026-automatic}. However, no systematic adaptation of MIP was carried out for Arabic up to the moment of writing this paper.

\subsection{Related Work}
Despite the absence of a systematic procedure for identifying metaphors in Arabic texts, we find multiple works that discuss metaphors in Arabic. \citet{alkhatib2016natural} discuss the definitions, usages and differences of metaphors across dialects. They compare Classical, Modern Standard, and Dialectal Arabic. The authors argue that machine translation approaches do not translate metaphors correctly.
\citet{alsiyat-piao-2020-metaphorical} show the importance of metaphor detection and interpretation in the task of sentiment analysis and later develop the Arabic Metaphor Corpus~\cite[AMC,][]{amc-corpus}, which contains 1,000 sentences extracted from book reviews annotated with metaphor spans and sentiment. However, no information on the procedure used to annotate the metaphors is mentioned.
\citet{magdy-etal-2024-gazelle} introduce a dataset for Arabic writing assistance, which includes 330 sentences with incorrect usage of metaphors or multi-word expressions (MWEs) paired with their corrections. However, the authors do not explain the process of collecting these sentences. 
\citet{zibin2025metaphor} conduct a study where they compare human performance in interpreting (Classical, Jordanian, and Emirati) Arabic metaphors to Large Langauge Models (LLMs), finding that it is challenging for LLMs to process culturally rich and context-driven language compared to humans. \citet{banou2025memphis} introduce a dataset of Arabic figurative speech, including metaphors and similes, but they are directly translated from English. 

Overall, we find a clear gap in work on systematically identifying metaphors in Arabic using a theoretically grounded procedure that has been applied to and validated across other languages.

\section{AraMIP: Annotation Framework}
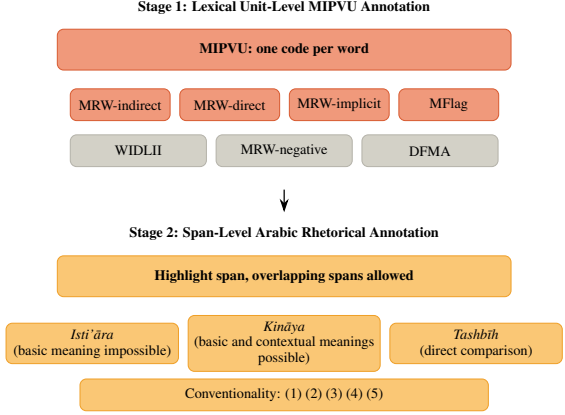
\begin{figure}[h]
\centering
\scalebox{0.60}{
\usetikzlibrary{arrows.meta}
\begin{tikzpicture}[
  boxa/.style={rectangle, rounded corners=4pt, draw=gray, fill=#1, minimum width=10cm, minimum height=0.9cm, align=center, font=\small},
  pill4/.style={rectangle, rounded corners=4pt, draw=gray, fill=#1, minimum width=2.2cm, minimum height=0.7cm, align=center, font=\footnotesize},
  pill3/.style={rectangle, rounded corners=4pt, draw=gray, fill=#1, minimum width=3.0cm, minimum height=0.7cm, align=center, font=\footnotesize},
  node distance=0.4cm
]

\definecolor{stageonefill}{RGB}{240,153,123}
\definecolor{stageonedraw}{RGB}{216,90,48}
\definecolor{stagetwofill}{RGB}{250,199,117}
\definecolor{stagetwodraw}{RGB}{239,159,39}
\definecolor{neufill}{RGB}{211,209,199}
\definecolor{neudraw}{RGB}{180,178,169}

\node[font=\small\bfseries] (s1title) {Stage 1: Lexical Unit-Level MIPVU Annotation};
\node[boxa=stageonefill, draw=stageonedraw, below=0.2cm of s1title] (s1)
  {\textbf{MIPVU: one code per word}};

\node[pill4=stageonefill, draw=stageonedraw, below=0.4cm of s1, xshift=-1.2cm] (b) {MRW-direct};
\node[pill4=stageonefill, draw=stageonedraw, left=0.2cm of b]   (a) {MRW-indirect};
\node[pill4=stageonefill, draw=stageonedraw, right=0.2cm of b]  (c) {MRW-implicit};
\node[pill4=stageonefill, draw=stageonedraw, right=0.2cm of c]  (d) {MFlag};

\node[pill3=neufill, draw=neudraw, below=0.3cm of b, xshift=1.2cm] (e) {MRW-negative};
\node[pill3=neufill, draw=neudraw, left=0.2cm of e]  (g) {WIDLII};
\node[pill3=neufill, draw=neudraw, right=0.2cm of e] (f) {DFMA};

\node[font=\small\bfseries, below=1.2cm of e] (s2title)
  {Stage 2: Span-Level Arabic Rhetorical Annotation};
\coordinate (s1bottom) at ([yshift=-5mm]e.south);
\coordinate (s2top) at ([yshift=2mm]s2title.north);
\draw[-{Stealth[length=3mm, width=2mm]}, thick, black]
  (s1bottom) -- (s1bottom |- s2top);

\node[boxa=stagetwofill, draw=stagetwodraw, below=0.2cm of s2title] (s2)
  {\textbf{Highlight span, overlapping spans allowed}};

\node[pill3=stagetwofill, draw=stagetwodraw, minimum width=3.8cm, below=0.4cm of s2] (kin)
  {\textit{Kināya}\\(basic and contextual meanings\\possible)};
\node[pill3=stagetwofill, draw=stagetwodraw, minimum width=3.8cm, left=0.2cm of kin] (isti)
  {\textit{Isti'āra}\\(basic meaning impossible)};
\node[pill3=stagetwofill, draw=stagetwodraw, minimum width=3.8cm, right=0.2cm of kin] (tash)
  {\textit{Tashbīh}\\(direct comparison)};

\node[pill3=stagetwofill, draw=stagetwodraw, minimum width=9cm, below=0.3cm of s2, yshift=-1.5cm] (conv)
  {Conventionality: (1) (2) (3) (4) (5)};

\end{tikzpicture}
}
\caption{The AraMIP two-stage annotation procedure.}
\label{fig:aramip}
\end{figure}

AraMIP is a two-stage annotation framework for Arabic metaphor and related figurative language annotation. AraMIP extends MIPVU with a second annotation layer based on the Arabic rhetorical tradition. The combined framework retains MIPVU’s core annotation categories and decision principles while providing additional information about the rhetorical nature of identified metaphorical expressions (see Figure~\ref{fig:aramip}).

\subsection{Stage~1: MIPVU Annotation}
In Stage 1, AraMIP adopts the MIPVU procedure and annotation categories. The MSA dictionary \emph{`\small{\<معجم اللغة العربية المعاصرة>}\normalsize'} \textit{Mu`jam al-Lugha al-`Arabiyya al-Mu`\=a\d{s}ira \footnote{The dictionary was accessed via the \url{https://www.almaany.com/} interface.}}~\citep{omar2008muajam} is used to determine the ``basic'' meaning of each lexical unit in turn. Metaphor-related words (MRWs) are identified where a lexical unit participates in a cross-domain relationship between a contextual meaning and a more basic meaning, whether expressed directly, indirectly, or through an associated metaphor signal.

Direct metaphors are identified through explicit cross-domain comparison, typically signaled by a lexical comparator such as a simile marker, whereas this is absent in indirect metaphors. Implicit metaphors are lexical units that are MRWs by virtue of a previously stated comparison. Thus, each lexical unit is annotated as one of:  indirect metaphor ``MRW-indirect'', direct metaphor ``MRW-direct'', implicit metaphor ``MRW-implicit'', metaphor flag ``MFlag'' (for simile markers), or not a MRW ``MRW-negative''. Other possible MIPVU annotations are \emph{When In Doubt, Leave It In} (``WIDLII'') for uncertain cases, and \emph{Discarded For Metaphor Analysis} (``DFMA'') for when the meaning of a lexical unit is impossible to discern.

\subsection{Stage~2: Arabic Rhetorical Classification}
The direct and indirect metaphors identified in Stage~1 generally correspond to three rhetorical devices in Arabic. Although the pattern is not invariable, direct metaphors often correspond to \textit{tashbīh} (simile), while indirect metaphors often correspond to \textit{isti‘āra} (metaphor) and \textit{kināya} (metonymy/indirect expression). Both the ``basic'' and ``contextual'' meanings are possible in \textit{kināya}, but not in \textit{isti‘āra}. AraMIP therefore introduces a second annotation layer applied to non MRW-negative lexical units.

In accordance with Arabic rhetorical tradition, Stage~2 uses the rhetorical construction as the unit of annotation rather than the lexical unit. Rather than annotating only the individual metaphor-related word, annotators identify the complete span of the rhetorical construction to which that word belongs. Unlike Stage~1, which operates at the lexical-unit level, Stage~2 permits multi-word spans and overlapping rhetorical constructions. MFlag lexical units are annotated as part of the \textit{tashbīh} span they introduce, rather than as an independent rhetorical category; MRW-implicit lexical units are not annotated in Stage~2.

While Stage~1 identifies conventionalized and novel metaphors alike, it does not record their degree of conventionality. In Stage~2, AraMIP additionally records the degree of conventionality of each rhetorical construction on a five-point scale.

The resulting corpus, therefore, contains both lexical unit-level metaphor annotations and span-level rhetorical annotations. Unlike previous MIPVU-based corpora, AraMIP supplements lexical-level metaphor annotation with span-level rhetorical classification grounded in the Arabic rhetorical tradition.

\section{Corpus Selection}
\label{sec: corpus_selection}

To test our proposed annotation framework, we select a collection of sentences to be annotated and analyzed in a pilot study. For simplicity, we only consider MSA texts, aiming to include different genres and domains to create a representative pool of MSA usage. We base this on VUAMC~\cite{mipvu}, in which the annotated sentences are sampled from the British National Corpus (BNC), thus seeking a similarly diverse MSA corpus that preferably contains additional annotation layers. 

We find BAREC-10M~\cite{elmadani-etal-2026-large} to be a suitable choice for our purpose. BAREC-10M contains 10 million words from a range of text genres, where each document is split by sentences and tokens and manually annotated for: \begin{enumerate*}[label=\arabic*.] \item \textbf{Domain} (\textit{Arts \& Humanities}; \textit{Social Sciences}; \textit{STEM}); \item \textbf{Readership group} (\textit{Foundational}; \textit{Advanced}; \textit{Specialized}); and \item \textbf{Text category} (\textit{Media \& Culture}; \textit{Religion \& Philosophy}; \textit{Literature, Art \& Music}; \textit{Academic}; \textit{Educational Materials, Language \& Linguistics}; \textit{Encyclopedic}). \end{enumerate*} Additional automatic annotations are also available, such as word-level and sentence-level readability scores and part-of-speech tags.

BAREC-10M contains around 514K sentences, from which we use 300 sentences to conduct our pilot study. We sample 50 random sentences from each text category, ensuring an equal distribution across domains and readership groups when possible. We also restrict sentences to lengths between 8 and 30 words, resulting in a corpus of 300 sentences and 5,277 words, with an average sentence length of 17.58 words.

\section{Annotation Pilot Study}


\subsection{Annotation Setup}

\paragraph{Annotators} Annotation was performed by six annotators: five native Arabic speakers representing Egyptian and Levantine dialects, and one L2 Arabic speaker. Collectively, the annotators had expertise in Arabic NLP, lexicography, and Arabic rhetorical and figurative language. Before the main annotation process, all annotators participated in a training session during which they jointly annotated and discussed a sample of 10 sentences to establish a shared understanding of the guidelines. 

\paragraph{Data} The 300 sampled sentences were randomly partitioned and assigned to the annotators while ensuring an approximately equal distribution across text domains. Each sentence was independently annotated by three different annotators.

\paragraph{Annotation Platform} We use INCEpTION~\citep{tubiblio106270}, a web-based annotation platform, to annotate both stages. Screenshots of the annotation interface are shown in Appendix~\ref{app:annotation-tool}.

\subsection{Inter-annotator Agreement}
We evaluate inter-annotator agreement (IAA) separately for each annotation stage. For Stage~1, where each word is assigned an MRW label, we report agreement at two levels. First, we measure agreement on whether a word is MRW-positive or -negative, regardless of its specific MRW type. Second, we measure agreement using the full set of MRW labels. For both settings, we compute Fleiss' $\kappa$ and report the average score, the weighted average score, and the agreement for each label. The results are shown in Table~\ref{tab:iaa_word}.

For Stage~2, where annotators identify spans and assign figurative language labels to them, agreement depends on both the span boundaries and the assigned label. We compute pairwise exact-match F1 between annotators and report the average across all annotator pairs. Exact-match F1 requires both the span boundaries and the label to match exactly. We also compute pairwise IoU-based\footnote{Intersection over Union} F1 and average the scores across all annotator pairs, using an IoU threshold of 0.5 to allow partial overlap between spans. Finally, we report Soft $\gamma$ \cite{10.1162/COLI_a_00227} under three settings: equal weight for span position and label, higher weight for the label, and higher weight for the span position. The results are shown in Table~\ref{tab:iaa_span}.

\begin{table}[h]
\small
    \centering
    \begin{tabular}{l|c}
    \toprule
\textbf{Measure} & \textbf{Fleiss'} $\boldsymbol{\kappa}$ \\
\midrule
  Binary MRW agreement   &  0.358\\
  Avg. MRW label agreement & 0.229 \\
  Weighted MRW label agreement  &   0.357\\
  \midrule
  \multicolumn{2}{c}{\textbf{\textit{Per-label agreement}}}\\
  \midrule
MRW-Negative &  0.360 \\
MRW-Flag &   0.499 \\
MRW-Direct & 0.292  \\
MRW-Indirect & 0.315   \\
MRW-Implicit & -0.001   \\
MRW-DFMA & 0.104   \\
MRW-WIDLII & 0.035  \\
  \bottomrule
    \end{tabular}
    \caption{Inter-annotator agreement for the word-level annotation layer, measured using Fleiss' $\kappa$.}
    \label{tab:iaa_word}
\end{table}

\begin{table}[h]
\small
    \centering
    \begin{tabular}{l|c}
    \toprule
\textbf{Measure} & \textbf{Score} \\
\midrule
  Soft $\boldsymbol{\gamma}$ (equal weighting)   &  0.477\\
  Soft $\boldsymbol{\gamma}$ (label-priority weighting)  & 0.442 \\
  Soft $\boldsymbol{\gamma}$ (position-priority weighting) &   0.440\\
  \midrule
Exact-match F$_1$ &  0.125 \\
IoU-based F$_1$ &   0.211 \\
  \bottomrule
    \end{tabular}
    \caption{Inter-annotator agreement for the span-level annotation layer.}
    \label{tab:iaa_span}
\end{table}







\subsection{Automatic Curation}
\begin{figure}[t]
    \centering
    \includegraphics[width=\linewidth]{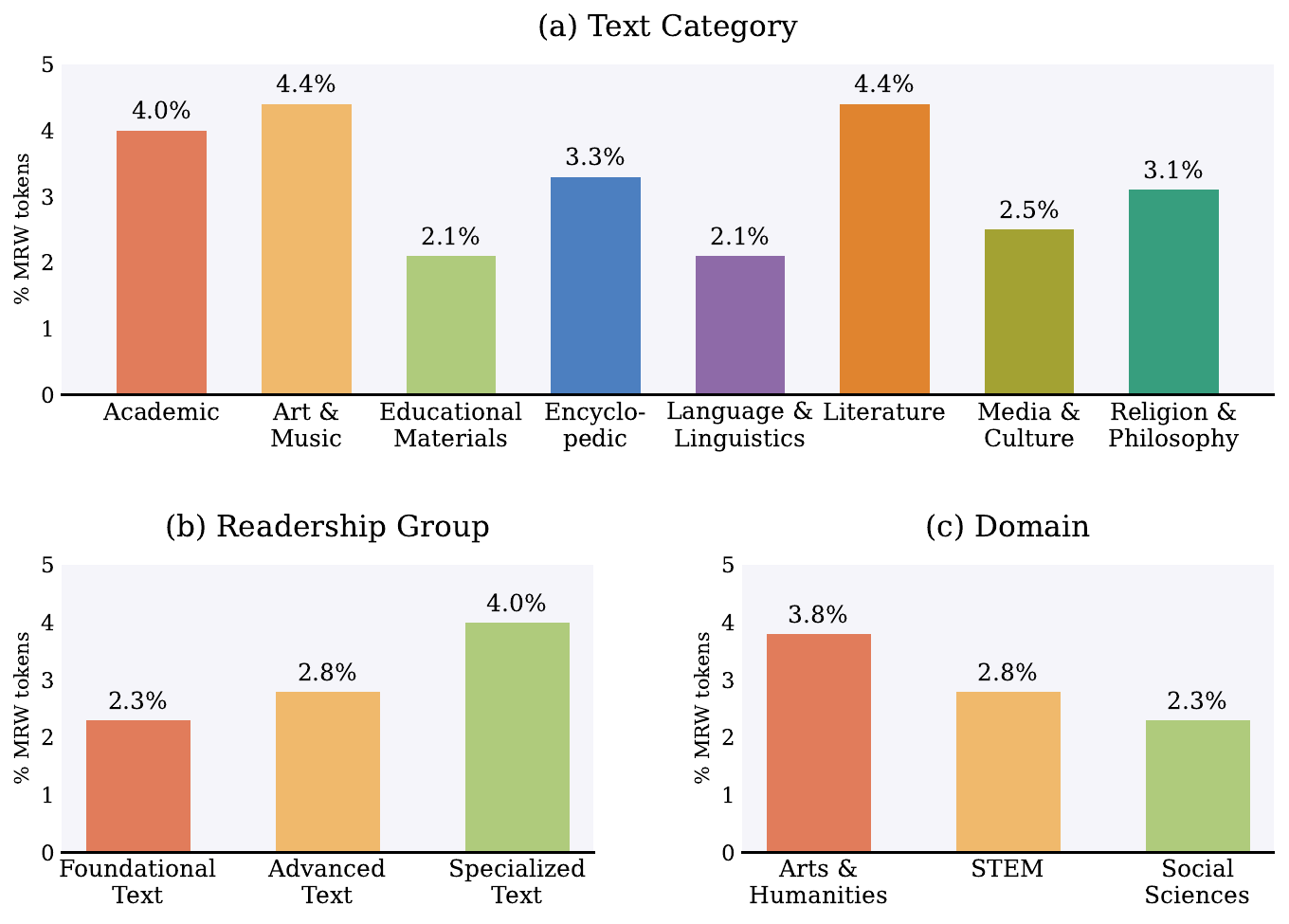}
    \caption{Percentage of MRWs across three corpus dimensions: text category, readership group, and domain.}
    \label{fig:mrw-distribution}
\end{figure}

We curate the annotations for both Stage~1 and Stage~2 for each word/sentence as follows: 

\paragraph{Stage~1} Each word was assigned an MRW code following a majority vote. For the curation process, we distinguish between three kinds of agreement cases: 1. \textit{Clear agreement:} same label given by all annotators, 2. \textit{Majority agreement:} two annotators agree on the label, and 3. \textit{Disagreement:} three different labels. The curated dataset contains 90.49\% \textit{clear agreement} cases, 9.08\% \textit{majority agreement} cases, and 0.44\% \textit{disagreement} cases. 
 To resolve the disagreement cases, we perform a second annotation round, in which three different annotators independently annotate each case. The final curation was then performed by considering the annotations from all six annotators. This process resolved 21 out of 23 disagreements, leaving two critical disagreements for discussion, in which the words were still annotated with three different labels each by two annotators.
The first disagreement concerned the word \emph{``\small{\<مفاجأة>}\normalsize'' - ``surprise''} in the sentence \emph{``\small{\<كان الرّجل مفاجأة لا تنتهي بالنسبة لي>}\normalsize'' - ``The man was a never-ending surprise to me''} which was annotated with the three labels \textit{MRW-Direct}, \textit{MRW-Indirect} and \textit{MRW-Negative} twice each. 
The second disagreement involved to the word \emph{``\small{\<عاده>}\normalsize'' - ``Habit or something one is accustomed to''} in the sentence \emph{``\small{\<فلهذا جفاهُ من كان عادهُ>}\normalsize'' - ``Thus the one to whom it was accustomed forsook it''} which was annotated with the three labels \textit{MRW-WIDLII}, \textit{MRW-Indirect} and \textit{MRW-Negative} twice each. 
The curation after the second round resulted in 90.49\% \textit{clear agreement} cases, 9.48\% \textit{majority agreement} cases (out of which 75.8\% are \textit{MRW-Negative}), and 0.04\% \textit{disagreement} cases. Table~\ref{tab:stage1|_ann_pre_curation} and Table~\ref{tab:stage1_stats} describe the annotated dataset before and after the curation, respectively. We additionally report percentages of MRW-positive instances per text category, readership group, and domain in Figure~\ref{fig:mrw-distribution}.

\begin{table}[!htp]
    \centering
    \small
    \begin{tabular}{l|ccc}
    \toprule
         \textbf{Label} & \textbf{A1} & \textbf{A2} & \textbf{A3} \\
         \midrule
        MRW-Negative & 4964 & 5055 & 5018\\
        MRW-Indirect & 266 & 152 & 240\\
        MRW-Direct & 7 & 37 & 8\\
        MRW-Flag & 6 & 8 & 4\\
        MRW-Implicit & 8 & 2 & 1\\
        MRW-Direct/Flag & 4 & 5 & 0\\
        MRW-WIDLII & 15 & 10 & 2\\
        MRW-DFMA & 7 & 8 & 4\\
         \bottomrule
    \end{tabular}
    \caption{Distribution of Stage~1 MRW labels assigned by the three different annotators: A1, A2, A3.}
    \label{tab:stage1|_ann_pre_curation}
\end{table}

\begin{table}[!htp]
\centering
\small
\begin{tabular}{l|cc}
\toprule
\textbf{Label} & \textbf{\#} & \textbf{\%} \\ \midrule
MRW-Negative       & 5106           & 96.76            \\
MRW-Indirect       & 152            & 2.88              \\
MRW-Direct         & 6              & 0.11              \\
MRW-Flag           & 4              & 0.08              \\
MRW-Direct, MRW-Flag   & 4              & 0.08              \\
MRW-DFMA           & 2              & 0.04             \\
MRW-WIDLII         & 1              & 0.02              \\
Disagreement   & 2              & 0.04              \\ \bottomrule
\end{tabular}
\caption{Distribution of Stage~1 labels in the curated dataset, showing the number (\#) and percentage (\%) of annotations for each label.}
    \label{tab:stage1_stats}
\end{table}

\paragraph{Stage~2} Span-level annotations were curated using a strict boundary-matching criterion. For each sentence, we compare the spans produced by the three annotators and retain only spans for which at least two annotators assigned the same label and the span boundaries either matched exactly or one span was fully contained within the other. Thus, cases of mere partial overlap were excluded, while minor boundary differences were accepted when one annotator selected a more specific subspan of another annotator's longer span. To curate conventionality, we average the score given by the three annotators (we report the lowest scores, i.e. least conventional metaphors, in Figure~\ref{fig:conv-scores}). Table~\ref{tab:stage2|_ann_pre_curation} and Table~\ref{tab:stage2_stats} describe the annotated dataset before and after the curation, respectively.

\begin{table}[!htp]
    \centering
    \small
    \begin{tabular}{l|ccc}
    \toprule
         \textbf{Label} & \textbf{A1} & \textbf{A2} & \textbf{A3} \\
         \midrule
        isti‘āra & 180 & 139 & 181\\
        kināya & 70 & 41 & 46\\
        tashbīh & 9 & 9 & 9\\
        isti‘āra,kināya & 1 & 0 & 0 \\
         \bottomrule
    \end{tabular}
    \caption{Distribution of annotated spans across Stage~2 labels assigned by the three different annotators: A1, A2, A3.}
    \label{tab:stage2|_ann_pre_curation}
\end{table}

\begin{table}[!htp]
\centering
\small
\begin{tabular}{l|cc}
\toprule
\textbf{Label} & \textbf{\#} & \textbf{Conventionality\(_{AVG}\)} \\ \midrule
isti‘āra      & 106           & 4.05             \\
kināya       & 22            & 3.67              \\
tashbīh        & 8              & 2.40              \\
\bottomrule
\end{tabular}
\caption{Distribution of curated spans by Stage~2 labels. (\#) denotes the number of curated spans, and (Conventionality\(_{AVG}\)) denotes the mean conventionality score for each label.}
    \label{tab:stage2_stats}
\end{table}

\section{Disagreement Analysis}

\begin{figure*}[t]
    \centering
    \includegraphics[width=\linewidth]{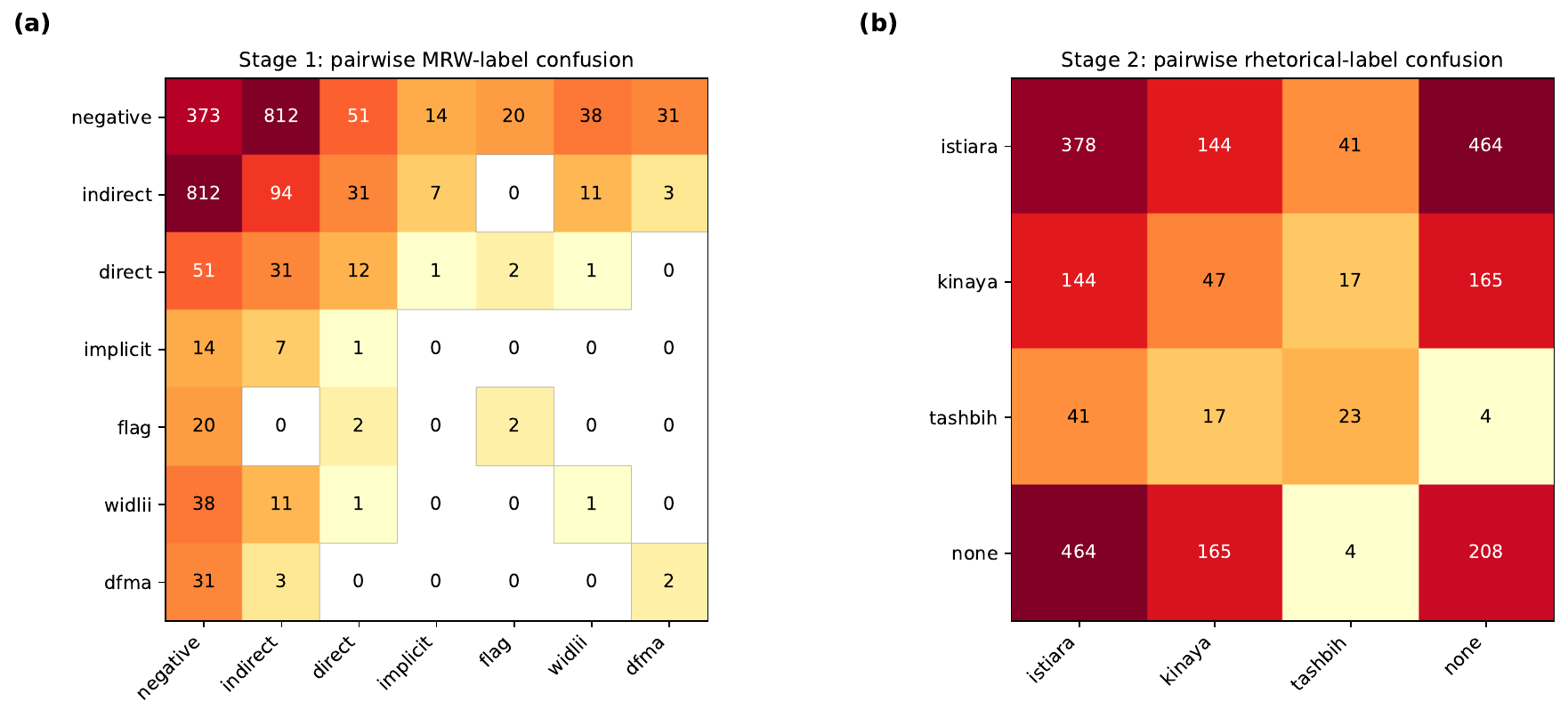}
    \caption{Pairwise label confusion matrices for (a) Stage 1 MRW labels and (b) Stage 2 rhetorical labels (\textit{none} denotes cases where an annotator did not assign a label).}
    \label{fig:disagreement}
\end{figure*}

To better understand the sources of annotator disagreements, we conduct a qualitative analysis of the disagreement cases from both annotation stages. We inspect the pairwise confusion patterns (Figure~\ref{fig:disagreement}) and manually examine the sentences with the highest degree of disagreement, identifying recurring linguistic and interpretive factors behind them. 

\paragraph{Stage~1 Disagreements} Of 502 tokens resolved by majority vote (479) or left as full disagreements (23), we see that the dominant confusion is MRW-indirect $\leftrightarrow$ MRW-negative (812 pairs), followed by MRW-direct $\leftrightarrow$ MRW-negative (51) and WIDLII $\leftrightarrow$ MRW-negative (38). Thus, we infer that the central difficulty is MRW-positive vs. -negative status rather than distinguishing among positive subtypes. Within MRW-positive, MRW-direct $\leftrightarrow$ MRW-indirect is the most frequent disagreement (31 pairs). We take a closer look at the 23 full disagreement tokens, which map out to 19 unique sentences sourced mainly from the Literature; Art \& Music category (36.8\%) and Arts \& Humanities domain (68.4\%), and identify five main categories of disagreement reasons:

\begin{enumerate}
    \item \textbf{Polysemy and domain-boundary ambiguity}: disagreement stems from deciding whether two senses of a word are distinct enough domains to count as metaphor, e.g. \emph{``\small{\<خطوة>}\normalsize'' - ``step''} as interpreted in the sentence \emph{``\small{\<ذكرت الكاتبة عدة [خطوات] في تقدير الذات>}\normalsize'' - ``The author mentioned several [steps] for self-appreciation''}.
    \item \textbf{Classical or poetic register}: disagreements most probably occur because sentences require specialized knowledge of Classical Arabic or external exegesis. For example, the token \emph{``\small{\<غررا>}\normalsize'' - ``blazes''} in \emph{``\small{\<رأى [غررا] لها حسن وماء>}\normalsize'' - ``it saw in it [blazes] of beauty and luster''}.
    \item \textbf{MWEs}: in some cases, figurativity attaches to a full expression rather than a single token, and annotators disagreed on which token(s) carry the figurative meaning. E.g., \emph{``\small{\<وضع الأمور في [نصابها]>}\normalsize'' - ``putting things in their proper place''}.
    \item \textbf{Dictionary gaps}: some tokens were absent from the reference dictionary and unattested elsewhere, causing annotation disagreement, e.g. \emph{``\small{\<العضلات [الهددية]>}\normalsize'' - ``the ciliary(?) muscles''}, which is possibly an error in the original corpus.
    \item \textbf{Cultural and religious influence}: some interpretations of metaphors depend on background knowledge or belief of an annotator, which in turn would cause disagreements, e.g. \emph{``\small{\<أبواب الله>}\normalsize'' - ``the doors of God''}, where a literal vs. figurative reading depends on the annotator's own religious view.
\end{enumerate}

\paragraph{Stage~2 Disagreements} Of 136 curated spans, 45 had full agreement, 91 were resolved by majority vote, and 4 spans had complete disagreement (excluded from the final curated set). The most frequent source of disagreement is between a labeled span and unlabeled for both isti‘āra and kināya, i.e., annotators disagreed on whether a span was figurative at all; tashbīh shows comparatively high agreement, likely due to its explicit comparator. Excluding unlabeled cases, the most common category of disagreement is isti‘āra vs. kināya (44 out of 55 cases), which centers on whether the span's meaning can happen literally or not, as in the example \emph{``\small{\<مَسَّ الْقَوْمَ قَرْحٌ>}\normalsize'' - ``a wound has touched the people''}.

\section{Discussion}

\subsection{Dictionary-related Challenges}
\paragraph{Coverage and the Modern/Classical boundary.} A recurring difficulty during annotation concerned the choice and coverage of the Arabic dictionary used to determine basic (non-contextual) word meanings. Although we deliberately selected a modern Arabic dictionary to match the contemporary nature of our corpus, the boundary between MSA and Classical Arabic vocabulary proved difficult to differentiate. Many words occurring in classical registers, including terms found in Quranic verses or traditional Arabic poetry, were absent from or inadequately treated in the dictionary. For example, the lexeme \emph{`\small{\<خصيلة>}\normalsize'} (a classical variant of \emph{`\small{\<خصلة>}\normalsize'}) was not included with its classical sense, meaning annotators could not reliably retrieve the basic meaning of such tokens. While the comprehensiveness constraints of any lexicographic resource are understandable, this created systematic gaps in the annotation workflow. Future guidelines should either specify a broader set of approved reference resources or provide explicit fallback procedures for cases of dictionary absence.

\paragraph{Morphological complexity and root-based lookup.} Arabic's rich morphological structure presents an additional challenge not addressed in the original MIPVU framework. Arabic dictionaries are organised by root, meaning that identifying the basic meaning of a morphologically derived form requires the annotator to locate the appropriate root entry and then determine whether the meaning of the derivative is predictable from, or has diverged from, that of the root. In practice, we observed a non-trivial mismatch between the primary sense listed under a root entry and the conventional meaning of a derived nominal or verbal form. MIPVU does not specify how to handle such cases, leaving annotators to exercise individual judgement. We consider this a key area for adaptation in future versions of AraMIP, which should provide explicit decision criteria for root-to-derivative meaning transfer.

\paragraph{Inconsistency in sense ordering.} Finally, although the dictionary we used states in its introduction that senses are ordered from most general to most specific, which is a convention that AraMI relies upon to identify the basic meaning of a lexical unit, we found this ordering to be inconsistently applied across entries in practice. This introduced further ambiguity into the sense selection step. Future work should either select resources with more reliable sense ordering, empirically validated if possible, or develop supplementary procedures that do not depend on lexicographic sense ordering alone.

\subsection{Distribution of Metaphor-Related Words}

The relatively low proportion of MRW-positive labels in the final corpus is expected given the composition of the sampled data. Our pilot sentences were drawn from a large-scale, domain-diverse corpus covering a wide range of MSA text types. The subset we used includes genres such as multiple-choice questions and scientific writing, which are predominantly denotative and thus inherently low in figurative density. Furthermore, our data is too small to ensure representativeness across the full range of MSA text types. These factors together limit the extent to which the distribution of annotation labels can be taken as indicative of broader figurative language use in Arabic. Larger-scale annotations covering a more balanced selection of domains will be necessary to characterize the figurative density of Arabic text more reliably.

\subsection{Inter-Annotator Agreement}
The IAA scores obtained in our pilot study fall below those reported for other MIPVU studies, including Fleiss' $\kappa$ 0.84 for the original VUAMC, averaged across six reliability tests conducted over two years~\citep{mipvu}. This reflects several practical and methodological challenges in annotating figurative language, particularly in a low-resource setting such as Arabic.

\paragraph{Impact of label imbalance.}
The annotated sample contains only a small number of MRW-positive instances, whereas the annotators agreed on the negative label for more than 90\% of the examples. Although this produces high observed agreement, it also increases the level of agreement expected by chance. Fleiss' $\kappa$ adjusts for this chance agreement. Therefore, because annotators are very likely to assign the dominant negative label, the resulting Fleiss' $\kappa$ score may remain low despite the high percentage of observed agreement.

\paragraph{Annotation experience and iterative refinement.} A key factor that contributes to disagreements is the limited number of annotation rounds. Due to time and resource constraints, annotators did not engage in the multiple cycles of annotation, discussion, and reconciliation that are usually recommended for developing new annotation guidelines~\cite{artstein2008inter}. This led to inconsistencies in the interpretations of certain annotation rules. A prominent example is the label combination for direct metaphors. According to MIPVU, direct metaphors are typically accompanied by a corresponding \emph{MFlag} label in the same sentence. However, a direct metaphor such as \emph{``\small{\<أحمد أسد في الملعب>}\normalsize'' - ``Ahmad is a lion on the field''} - lacks an \emph{MFlag} element. Whether this should be annotated as \emph{MRW-direct} or \emph{MRW-indirect} was not interpreted consistently by the annotators, and would likely be resolved through further discussion rounds.

\paragraph{Context availability and interpretation.} Unlike the original MIPVU, which annotated sentences with access to their full surrounding discourse, our annotation setup did not provide annotators with context material. When contextual information was needed, annotators were permitted to independently consult external resources such as online exegetical works. While this allowed for a more informed interpretation of individual tokens, it introduced a source of systematic variance: different annotators may have consulted different resources or weighted conflicting interpretations differently. Future work building on AraMIP should supply all annotators with a shared, fixed set of contextual materials and restrict independent look-ups to ensure that contextual knowledge does not become a confounding variable.


\paragraph{Annotator diversity.} The annotation team consisted of six annotators with varied disciplinary backgrounds. This interdisciplinary composition is a strength of the study in terms of coverage and perspective, but it also introduced variability in how figurative language was conceptualized and operationalized. Furthermore, while five annotators were native Arabic speakers, they represented four distinct geographic varieties of Arabic, and one annotator was an advanced L2 speaker specializing in Arabic linguistics. Dialectal variation and differences in rhetorical intuition may have influenced judgments. 
To assess whether this diversity in disciplines and dialects systematically affected annotation agreement, we conducted an analysis comparing the IAA across annotator backgrounds. The results showed that annotators, including the L2 speaker, had very similar average IAA scores, regardless of their disciplines or dialect backgrounds. In particular, pairs of annotators with similar or different dialects did not show any consistent pattern of higher or lower agreement. Detailed results of this analysis are provided in Appendix~\ref{sec:iaa_anal}.

\section{Conclusion}
In this work, we introduced AraMIP, the first systematic, two-stage procedure for identifying and classifying metaphor-related expressions in Arabic text. AraMIP extends MIPVU's lexical-unit-level metaphor identification with a second annotation layer that maps metaphor-related words onto the Arabic rhetorical tradition (\textit{isti‘āra}, \textit{kināya}, and \textit{tashbīh}), additionally recording the conventionality of each rhetorical construction on a five-point scale. We applied AraMIP to a pilot corpus of 300 sentences (5277 words) sampled from BAREC-10M across a balanced range of domains, readership levels, and text categories. Our IAA analysis shows that annotating Arabic figurative language reliably is more difficult than reported for several other MIPVU adaptations. We use this pilot study to identify concrete, addressable sources of this difficulty: a single annotation round without iterative guideline refinement, inconsistent dictionary coverage across the MSA/Classical Arabic boundary, unresolved root-to-derivative meaning transfer within Arabic's morphological system, inconsistent sense ordering within the lexicographic resource used, and uncontrolled access to contextual and exegetical materials during annotation. These findings constitute a diagnostic roadmap for future iterations of AraMIP, which should incorporate disagreement-informed guideline revisions, standardized contextual materials, and explicit root-based sense-lookup criteria. We release our pilot corpus and annotation guidelines to support this future work, scaling to larger and more representative corpora, extending coverage to dialectal Arabic, and training computational models for Arabic metaphor detection.

\section*{Limitations}

This work has several limitations. First, the pilot corpus of 300 sentences (5,277 words) is too small to reliably characterize metaphor use across MSA. Second, AraMIP was applied only to MSA text, leaving open how the guidelines generalize to dialectal Arabic. Third, reliance on a single modern dictionary introduces gaps in Classical Arabic coverage, inconsistencies in sense ordering, and unresolved cases of root-to-derivative meaning transfer. Fourth, annotation guidelines were not iteratively refined through multiple rounds of discussion before the main annotation phase, and annotators were not provided standardized contextual or exegetical materials, independently consulting external resources such as Tafaseer when needed. 

\section*{Acknowledgments}
This research has been funded by the \href{https://www.dfg.de/}{Deutsche Forschungsgemeinschaft} (DFG, German Research Foundation) -- \href{https://gepris.dfg.de/gepris/projekt/512393437}{CRC-1646, project no. 512393437}, project \href{https://gepris.dfg.de/gepris/projekt/537362555}{A05} and \href{https://gepris.dfg.de/gepris/projekt/537416633}{B02}.
\bibliography{custom}
\appendix
\section{Risks and Ethical Considerations}

We do not believe that there are significant risks associated with this work, as we annotate existing data without content that might be perceived as hurtful. 

\section{MIPVU Procedure}
\label{sec: apx_mipvu}
\begin{figure*}[!htp]
    \centering
    \includegraphics[width=1\linewidth]{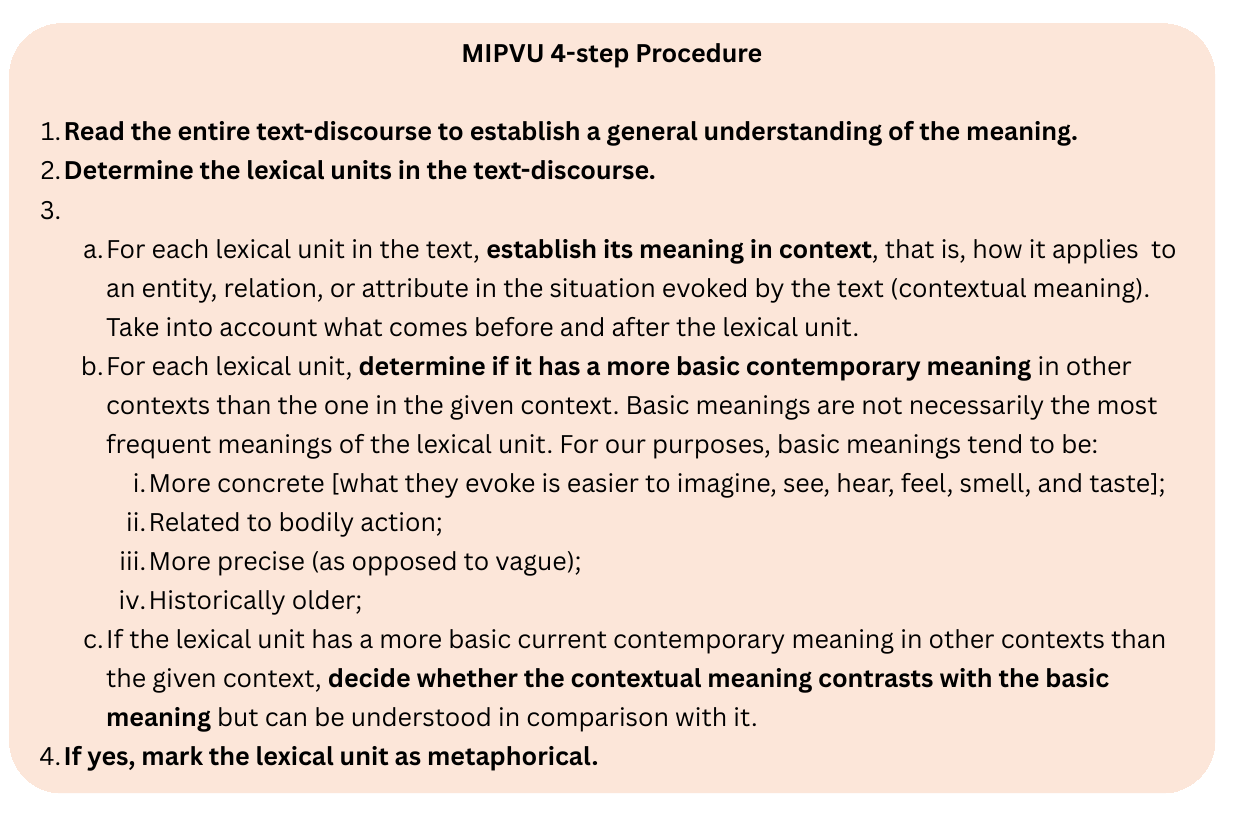}
    \caption{The Metaphor identification procedure of~\citet{mipvu}.}
    \label{fig:mipvu}
\end{figure*}

\citet{mipvu} propose a standard procedure of identifying metaphors in written and spoken discourse. This procedure is presented in Figure~\ref{fig:mipvu}.

\section{Annotation Tool}
\label{app:annotation-tool}
\begin{figure*}
    \centering
    \includegraphics[width=0.9\linewidth]{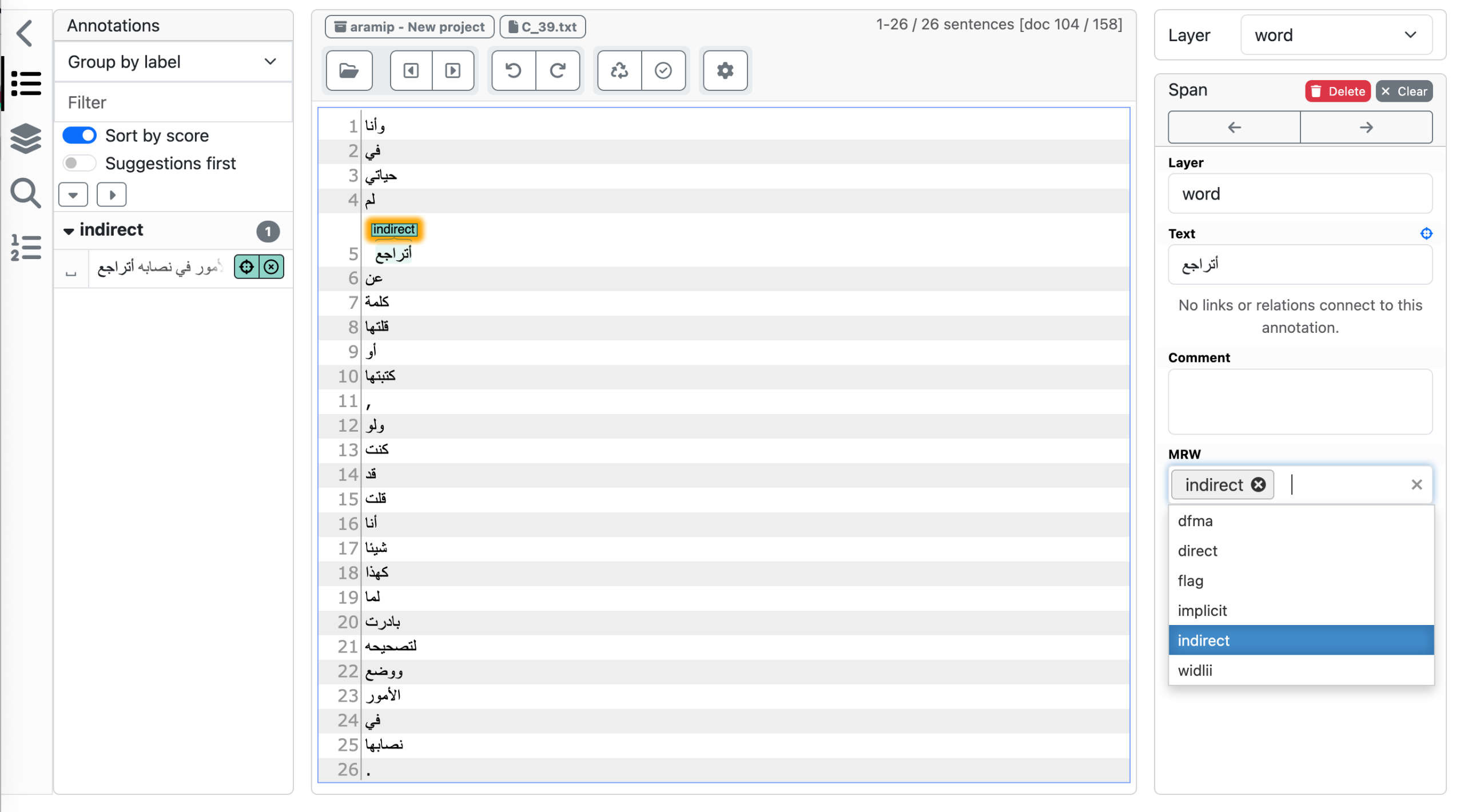}
    \caption{The INCEpTION annotation interface showing Stage 1 word-level MIPVU annotation, where each word is assigned one of seven codes.
    }
    \label{fig:app_level1}
\end{figure*}

\begin{figure*}
    \centering
    \includegraphics[width=0.9\linewidth]{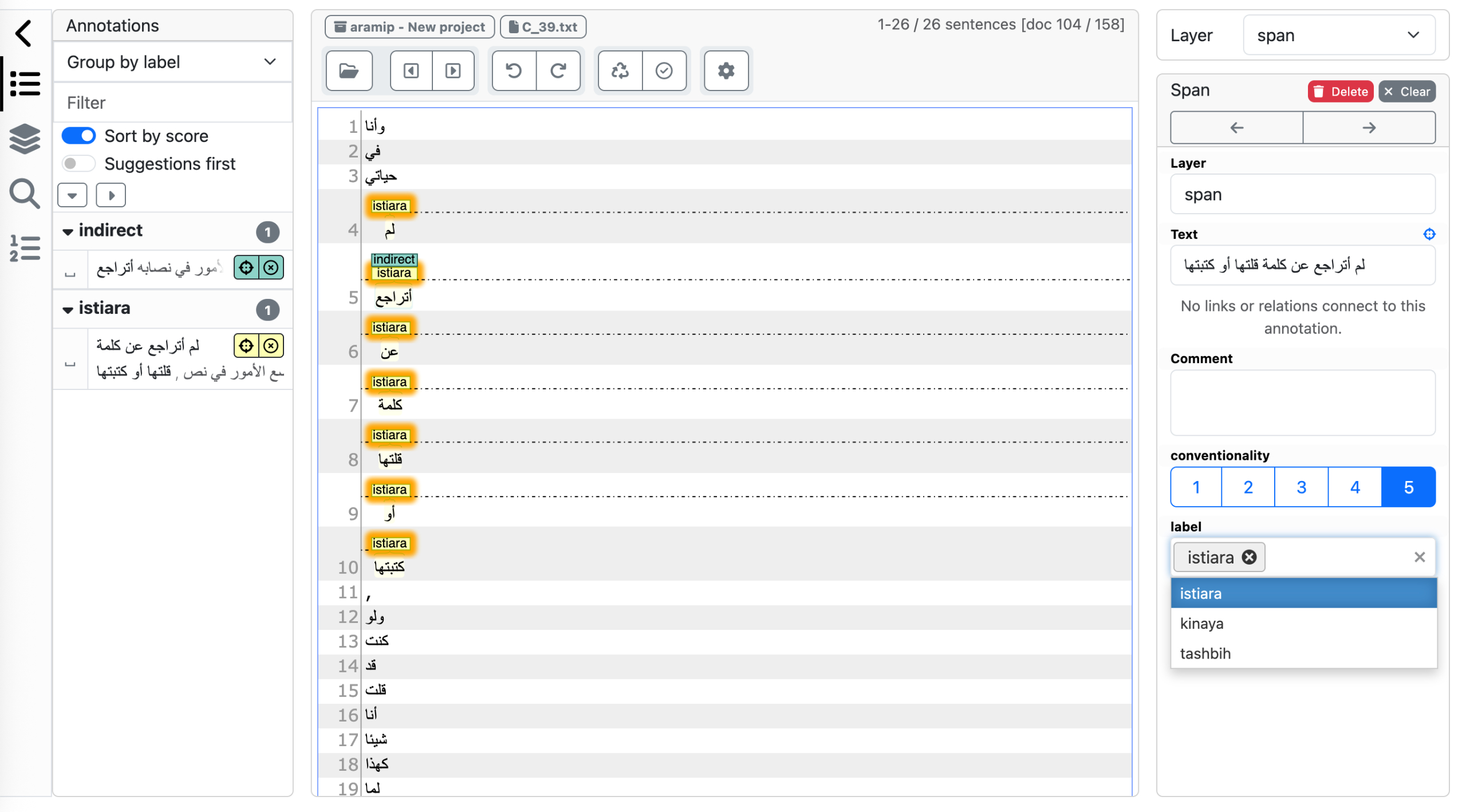}
    \caption{The INCEpTION annotation interface showing Stage 2 span-level rhetorical annotation, where annotators highlight spans and assign a rhetorical label (Isti'āra, Kināya, or Tashbīh) along with a conventionality rating.
    }
    \label{fig:app_level2}
\end{figure*}

The stage 1 and stage 2 annotations of an example sentence using the INCEpTION interface are presented in Figures \ref{fig:app_level1} and \ref{fig:app_level2}, respectively.

\begin{figure*}
    \centering
    \includegraphics[width=\linewidth]{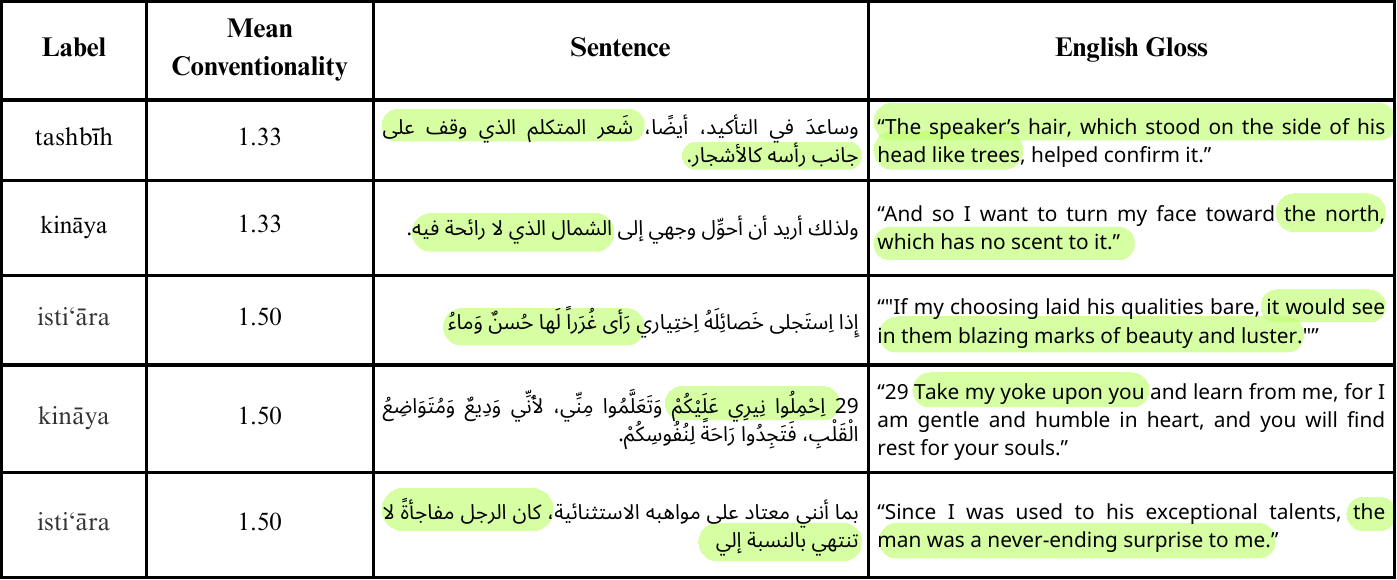}
    \caption{Five stage 2 span annotations with the lowest mean conventionality scores (on a scale of 1-5). The annotated span shown here is the longest span between the annotations.}
    \label{fig:conv-scores}
\end{figure*}

\section{Scientific Artifacts}
\label{app:scientific-artifacts}

In our work, we mainly used scientific artifacts in the form of publicly available datasets (Creative Commons Attribution Share Alike 4.0 International) and publicly available Python modules.

\section{Use of AI Assistants}

AI assistants were used during manuscript preparation for specific linguistic reformulation
to refine clarity and style, and to assist with code writing.

\section{License}
The AraMIP annotation guidelines and the resulting corpus are released under the Creative Commons Attribution-NonCommercial 4.0 International License (CC BY-NC 4.0).

\section{Recruitment Consent}
This study used the six authors as annotators. A training session on the use of AraMIP was provided before annotations were performed. This included practice with a training dataset. Annotators were instructed to follow the AraMIP annotation guidelines to annotate Arabic sentences from the study corpus. Annotators were recruited on the basis of their ability to perform the annotation task. They freely consented to participate in the annotation, and did not receive any financial compensation.

\section{IAA Results Across Disciplines and Dialects}
\label{sec:iaa_anal}
\begin{figure*}[h]
    \centering
    \includegraphics[width=0.8\linewidth]{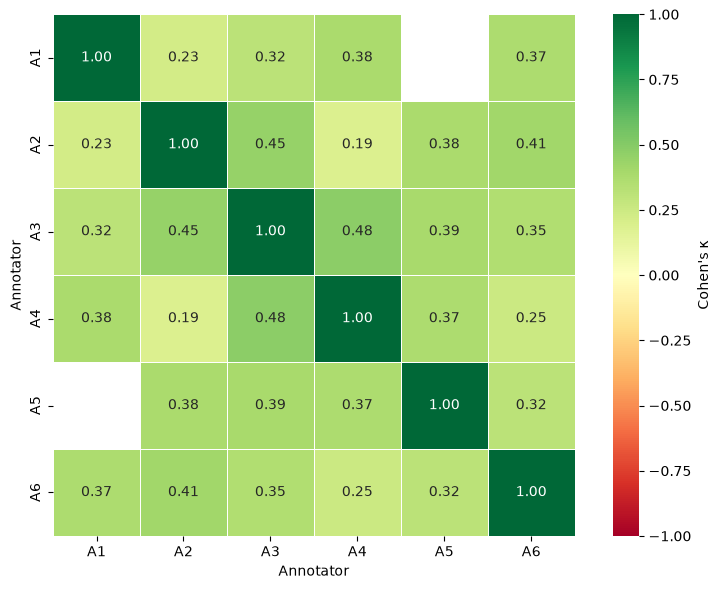}
    \caption{Pairwise Cohen's $\kappa$ agreement.}
    \label{fig:iaa_pair}
\end{figure*}

Table~\ref{tab:iaa_anal} shows details about the annotators' disciplinary and dialect backgrounds, along with their average pairwise Cohen's $\kappa$ agreement with the other annotators. Figure~\ref{fig:iaa_pair} shows the individual pairwise Cohen's $\kappa$ scores.

\begin{table*}[h]
    \centering
    \small
    \begin{tabular}{l|ccccc}
    \toprule
    \textbf{Annotaor}  &  \textbf{Arabic skills} & \textbf{Dialect} & \textbf{Gender} & \textbf{Expertise} &  \textbf{ Cohen's $\kappa$} \\
    \midrule
       A1  &  L1 & Syrian & M & Computer Science & 0.321 \\
        A2 & L1 & Palestinian \& Syrian & F & NLP & 0.330 \\
        A3 & L1 & Palestinian & F & NLP & 0.398 \\
        A4 & L1 & Palestinian \& Syrian & F & Cognitive Science & 0.333 \\
        A5 & L1 & Egyptian & M & Computer Science & 0.364 \\
        A6 & C1/2 level & Jordanian & M & Arabic rhetorical and figurative language
 & 0.340 \\
 \bottomrule
    \end{tabular}
    \caption{Annotators' disciplinary and dialect backgrounds, along with average pairwise Cohen's $\kappa$ scores.}
    \label{tab:iaa_anal}
\end{table*}

\end{document}